\pdfoutput=1
\documentclass[11pt]{article}

\usepackage{acl}
\usepackage{times}
\usepackage{latexsym}

\usepackage[T1]{fontenc}
\usepackage[utf8]{inputenc}

\usepackage{microtype}

\usepackage{inconsolata}

\usepackage{hyperref}
\usepackage{url}

\usepackage{algorithm}
\usepackage{algorithmic}
\usepackage[utf8]{inputenc} % allow utf-8 input
\usepackage[T1]{fontenc}    % use 8-bit T1 fonts
\usepackage{url}            % simple URL typesetting
\usepackage{booktabs}       % professional-quality tables
\usepackage{amsfonts}       % blackboard math symbols
\usepackage{nicefrac}       % compact symbols for 1/2, etc.
\usepackage{microtype}      % microtypography
\usepackage{bm}
\usepackage{enumitem}
\usepackage{graphicx}
\usepackage{amsmath}
\usepackage{subfigure}
\usepackage{amssymb}
\usepackage{caption}

\usepackage{comment}
\usepackage{xcolor}         % colors
\usepackage{centernot}
\newcommand{\notiff}{%
  \mathrel{{\ooalign{\hidewidth$\not\phantom{"}$\hidewidth\cr$\iff$}}}}
\usepackage{array}
\usepackage{multirow}
\usepackage{wrapfig,lipsum,booktabs}
\usepackage{makecell}
\usepackage{pifont}% http://ctan.org/pkg/pifont

\newcommand{\promptgpt}[1]{\texttt{``#1''}}

\usepackage{tabularx}

\usepackage{newfloat}
\usepackage{listings}

\title{\texttt{SCE}: Scalable Consistency Ensembles Make Blackbox Large Language Model Generation More Reliable}

\newcommand{\methodName}{\textsc{Sce}}
\newcommand{\methodNamenew}{\textsc{Sce}}
\newcommand{\saccheck}{\textsc{Sce-Check}}
\newcommand{\sacsummary}{\textsc{Sce-Fusion}}
\newcommand{\yopo}{\textsc{Yopo}}

\author{Jiaxin Zhang$^{1,2}$\thanks{Correspondence to jiaxin\_zhang@intuit.com.},  Zhuohang Li$^3$,  Wendi Cui $^1$,  Kamalika Das$^{1,2}$, \\ \bf{Bradley Malin$^{3,4}$, Sricharan Kumar$^{1,2}$}
\\ $^1$Intuit  $^2$Intuit AI Research  $^3$Vanderbilt University  $^4$Vanderbilt University Medical Center\\
}

\begin{document}
\ifdefined\pdfoutput
  \maketitle
\fi

\begin{abstract}
Large language models (LLMs) have demonstrated remarkable performance, yet their diverse strengths and weaknesses prevent any single LLM from achieving dominance across all tasks. Ensembling multiple LLMs is a promising approach to generate reliable responses but conventional ensembling frameworks suffer from high computational overheads and are thus not suited for LLMs. This work introduces Scalable Consistency Ensemble (\methodName), an efficient framework for ensembling LLMs by prompting consistent outputs. 
The \methodName~framework systematically evaluates and integrates outputs to produce a cohesive result through two core components: \saccheck,  a mechanism that gauges the consistency between response pairs via semantic equivalence; and \sacsummary, which adeptly merges the highest-ranked consistent responses from \saccheck, to optimize collective strengths and mitigating potential weaknesses. To improve the scalability with multiple inference queries, we further propose ``{\em You Only Prompt Once}'' (\yopo), a novel technique that reduces the inference complexity of pairwise comparison from quadratic to constant time. We perform extensive empirical evaluations on diverse benchmark datasets to demonstrate \methodName's effectiveness. Notably, the \saccheck~component outperforms conventional baselines with enhanced performance and a significant reduction in computational overhead.
 
\end{abstract}

\section{Introduction}
Large language models (LLMs) have demonstrated exceptional adaptability across diverse natural language tasks that require the generation of open-ended responses based on user prompt comprehension \cite{zhao2023survey, ski2024zhang}. Open-source LLMs, such as LLaMA \cite{touvron2023llama}, Pythia \cite{biderman2023pythia}, and Flan-T5 \cite{chung2022scaling} provide a flexible capability to fine-tune these models on custom datasets, enabling the development of efficient LLMs, including Alpaca \cite{alpaca}, Vicuna \cite{chiang2023vicuna}, OpenAssistent \cite{kopf2023openassistant}, Falcon \cite{refinedweb}, Dolly \cite{DatabricksBlog2023DollyV2}, etc. However, these smaller models struggle to achieve consistent performance due to limited generative capabilities. By contrast, prominent LLMs, such as ChatGPT, GPT-4 \cite{OpenAI2023GPT4TR}, and PaLM 2 \cite{anil2023palm}, have shown impressive and superior performance compared with open-source models. Although these closed-source LLMs are widely adopted in research and industry, they often tend to produce exceedingly confident, yet erroneous, assertions, referred to as \textit{hallucinations}. These hallucinated responses hinder the generation of reliable and robust responses and significantly impede the applicability of LLMs to real-world applications, especially in high-stakes areas, such as healthcare, criminal justice, and social services. 

\begin{figure*}[h!]
\centering
    \includegraphics[width=0.98\textwidth]{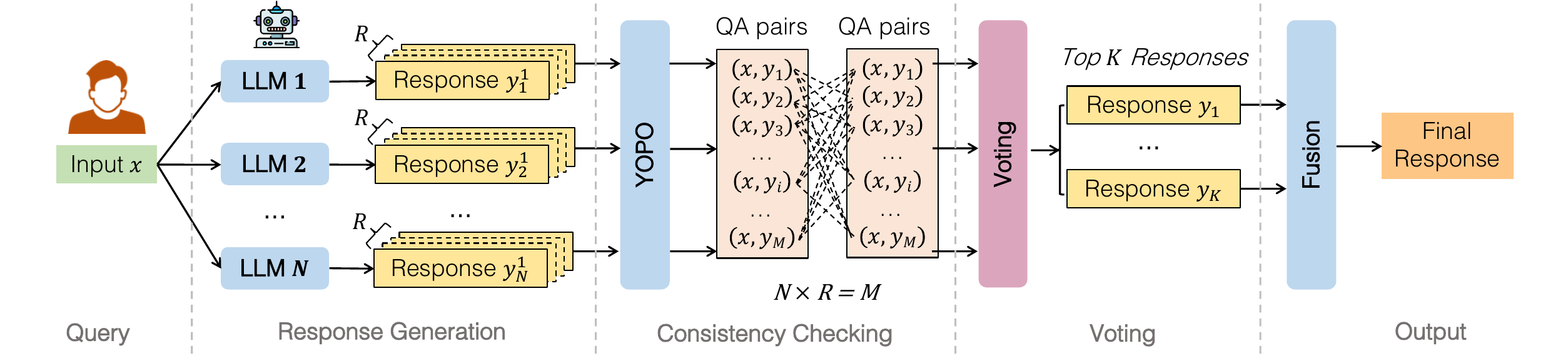} 
    \vspace{-5pt}
    \caption{Overview of the scalable consistency ensemble method, \methodName, towards reliable response generation.}
    \label{fig:overview_method}
    \vspace{-3mm}
\end{figure*}

These state-of-the-art LLMs, ChatGPT, GPT-4, and PaLM 2, exhibit variations in their strengths and weaknesses due to differences in architectures, model size, training data, and instruction fine-tuning, making them complementary to each other. Many existing model evaluation studies show that the performance of these models for different tasks can significantly vary and there is no single closed-source LLM that dominates all competitions. This naturally raises several practical questions: 

(1) {\em how to select an LLM given a specific task and user query} and (2) {\em how to identify a better response from outputs drawn from different LLMs}.
Yet, the decision does not need to be the selection of a single model. Our key insight is to {\em ensemble these LLMs to generate consistently reliable responses}. Considering the diverse advantages and limitations of LLMs, there is an opportunity to take advantage of their complementarity and build an ensemble approach that harnesses their complementary potentials, leading to enhanced reliability, robustness, accuracy, and generalization. By integrating their individual contributions, the misinformation, hallucination, and uncertainties in individual LLMs can be well-mitigated, resulting in outputs that are better aligned with human preferences.

Our idea of LLM ensemble is motivated by {\em self-consistency} in Chain of Thought (CoT) \cite{wang2022self} which selects the most consistent responses from a diverse set of sampled outputs, instead of only taking the one with the highest confidence. Self-consistency leverages intuition \cite{huang2022large} that a complex problem typically permits multiple different ways of thinking leading to its unique truthful answer, and can be naturally extended to open-text generation when a good metric of consistency is defined between multiple generated responses from LLMs. A recent study, selfcheckGPT \cite{manakul2023selfcheckgpt}, was proposed as an extension of self-consistency for hallucination detection which is built based on the essential assumption that if an LLM has knowledge of a given concept, sampled outputs are likely to be consistent; conversely, for hallucinated facts, sampled responses are likely to diverge and contradict one another. 

Building upon this intuition, Sac\textsuperscript{3}\cite{zhang2023sac} proposes a reliable hallucination detection strategy by introducing question-level and model-level perturbations incorporating checking semantic consistency instead of similarity metrics. However, pairwise consistency check used in selfcheckGPT and Sac$^3$ scales poorly in practice as the number of possible pairs grows quadratically with the increasing of responses. Moreover, both methods can only detect if a specific response is hallucinated or not but cannot provide an alternatively reliable response by utilizing sampled responses.

\paragraph{Proposed Approach.} We propose a scalable consistency ensemble framework, \methodName, to provide reliable performance by mixing the outputs of multiple prominent LLMs, as highlighted in Fig. \ref{fig:overview_method}.  

Specifically, our \methodName~consists of two core components: \saccheck, which compares output responses from multiple LLMs and checks the consistency among each other; and \sacsummary, which produces the final output via the most consistent votes.

For \saccheck, we first define a consistency metric to efficiently discern subtle differences between two LLM responses by prompting LLMs to return a binary score and experimentally validate that such a prompting approach performs more reliably and accurately. However, the native pairwise consistency checking suffers limited scalability. We further introduce a new strategy, named ``You Only Prompt Once'' (\yopo), which combines all responses in a single long prompt where the pairwise consistency can be checked via only one model inference, while significantly reducing the time complexity from $\mathcal{O}(N^2)$ to $\mathcal{O}(1)$.

For a specific response $R$, we check how many other responses are semantically equivalent to $R$ and count the number of consistent votes $\mathcal{V}_R$. Then we infer a ranking of the $N$ outputs for the given input and employ the top-ranked responses with the most votes as the final result. 
 
This approach may constrain the potential to generate even better outputs than the existing candidates. To investigate this possibility, we propose, \sacsummary, an additional component to fuse the top-ranked candidates and generate a comprehensive output via summarization. Our objective is to capitalize on the strengths of the top consistent candidates while mitigating their weaknesses. 

\begin{figure}[h!]
\centering
\includegraphics[width=0.48\textwidth]{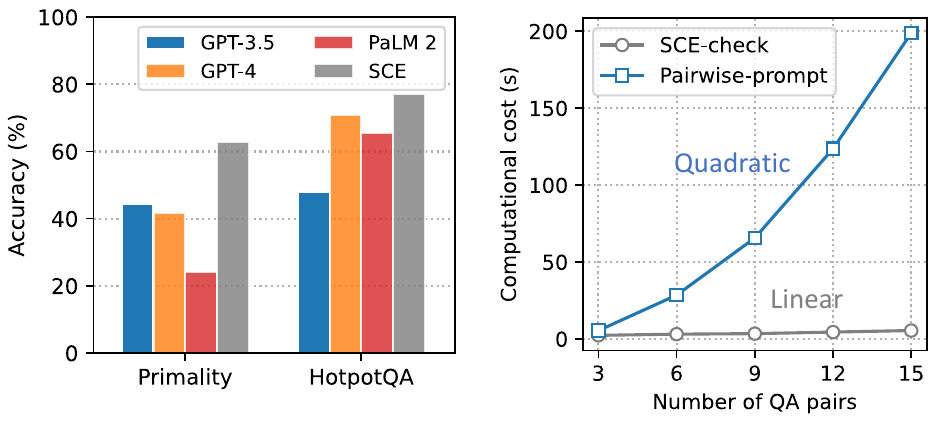}
\vspace{-5mm}
\caption{(Left) Truthfulness accuracy compared with a single fixed model; (Right) Computational cost of \saccheck~compared to the pairwise-prompt method.}
\label{fig:sce}
\end{figure}

In summary, our key contributions to this work
are threefold: (1) We propose a novel framework \methodName~tailored for generic blackbox LLM ensembles, leading to better and reliable responses compared to a single model; (2) We introduce a scalable strategy \yopo~for checking semantic consistency, reducing inference complexity from quadratic to consistent time; (3) We demonstrate superior performance in consistency check and response truthfulness through comprehensive experimental studies on multiple benchmark datasets across different LLMs, as shown in Fig. \ref{fig:sce}.    

\section{Preliminaries}
\subsection{Problem Formulation}
Given an input $x$ and a set of LLM candidates, $\mathbb{M} = \{ \mathcal{M}_1,..., \mathcal{M}_N \}$, we can generate $R$ candidate outputs by querying $x$ with each model $\mathcal{M}_k$. We denote the output responses as $\mathbb{Y} = \{y_1^1, y_1^2,...,y_1^R; y_2^1, y_2^2,...,y_2^R; ..., y_N^1, y_N^2,...,y_N^R \}$ where $y_k^r (1 \le k \le N, 1 \le r \le R)$ denotes the $r$-th sample drawn from the $k$-th model. Hence, there are $M = N\times R$ output responses in total for a specific input $x$. 

One may choose a single model, $\mathcal{M}_k (1 \le k \le N) \in \mathbb{M}$ to make an inference about $x$ and produce output $y_k = \mathcal{M}_k(x)$. This is only reasonable if $\mathcal{M}_k$ consistently demonstrates superior performance on all queried examples. However, relying on a pre-selected model may result in sub-optimal performance, as the $N$ models are likely to possess different strengths and weaknesses in various domains and tasks, meaning that the optimal model choice for different $x$ may not always be the same. 

Our objective is therefore to develop an ensemble approach that produces consistently better and reliable responses $y$ for different input $x$ by maximizing the {\em semantic consistency} between $y$ and the reference response $y^*$. We thus define a semantic consistency score as:
\begin{equation}
    \mathcal{S}(y^*, y | x) = \left\{\begin{matrix}
1.0, \quad  y \iff y^* \\
0.0, \quad  y \notiff y^*
\end{matrix}\right. \label{eq:score}
\end{equation}
where $\mathcal{S}(y^{*}, y | x) = 1.0$ if $y$ and $y^*$ are semantically equivalent and $\mathcal{S}(y^{*}, y | x) = 0.0$ otherwise. In many cases, the reference is unknown so this score metric $\mathcal{S}(y_i, y_j | x)$ can also be used to check semantic consistency in a generic response pair $(y_i, y_j), i \neq j$ where $y_i$ and $y_j$ are responses by either calling different models ({\em cross-evaluations}), 
\begin{equation}
    y_i = \mathcal{M}_{i}(x), \quad  y_j = \mathcal{M}_{j}(x), \quad \mathcal{M}_{i} \neq \mathcal{M}_{j}
\end{equation}
or calling the same model multiple times ({\em self-evaluations}). Unlike using a single fixed model or a randomly selected model, we propose an ensemble strategy to merge all responses from LLM self-evaluations and cross-evaluations together via $\hat{y} = \Omega(\mathbb{Y})$ where $\Omega(\cdot)$ is an Ensemble Operator built upon semantic consistency checking.

\subsection{Ensembles of Blackbox LLMs}
A drawback of the previous ``grey-box'' or ``white-box'' methods is that they require the token-level probabilities of the output for response clustering \cite{kuhn2023semantic}, while for prominent LLMs (e.g., GPT-3.5, GPT-4, PaLM 2, and Claude 2) that are only available through limited API calls, such token-level information might not be available. In contrast, we focus on designing a ``blackbox'' method that remains applicable even when only text-based responses are available from the LLM, i.e., we only need access to the LLM responses $y$ given the input query $x$ so our method works for any close-source or open-source LLMs. 

There are two primary methods for ensembling responses from blackbox LLMs: selection-based and generation-based. Selection-based methods compare candidates in the set $\mathbb{Y}$, selecting the top-ranked candidate as the final output $\hat{y}$ which implies that $\hat{y} \in \mathbb{Y}$. Due to the inherent nature of selection and the limited solution space, the performance of selection-based methods is bounded by the $N$ candidates being considered. Conversely, generation-based methods focus on fusing top-ranked consistent candidates from $\mathbb{Y}$ to produce an unseen response as the final output $\hat{y}$.

\section{Scalable Consistency Ensembles (\methodName): \\ A Novel Framework}
We propose a novel framework, \methodName, for ensembling blackbox LLMs through scalable semantic consistency check. This framework consists of two core components: a consistency check module, \saccheck, and a generative summarization module, \sacsummary. The \saccheck~module learns to compare consistency among all pairs of candidates and subsequently ranks the list of candidates according to consistency votes. We then select the top-ranked candidates, concatenate them with the input $x$, and construct the input sequence for the \sacsummary~module which ultimately generates the final output to serve users.

\subsection{Scalable Semantic Consistency Check} 
The proposed semantic consistency check approach consists of the following three key steps: 

\paragraph{Step 1: Response Generation.} Given an input query $x$, \saccheck~operates by drawing $R$ stochastic response samples $\{y_k^1, y_k^2,...,y_k^R\}$ from each LLM $\mathcal{M}_k, k = 1,...,N$, followed by measuring the consistency across the sampled responses. To better measure the consistency, we propose to check the semantic equivalence of two input-output pairs, by combining the input {\em question} $x$ with corresponding output {\em answer} $y_k^r$ as a question-answer (QA) pair, $(x, y_k^r)$. Then we combine all QA pairs $(x, y_k^r)$ to construct a QA pair matrix: 

\begin{equation}
\setlength{\abovedisplayskip}{5pt}
\setlength{\belowdisplayskip}{5pt}
\mathbb{QA}_{\textup{pair}} = 
\begin{bmatrix}
(x, y_1^1) & ...  &(x, y_1^R) \\ 
... &...  &... \\ 
(x, y_N^1) & ...  &(x, y_N^R))
\end{bmatrix},
\label{eq:matrix}
\end{equation}
where $M = N \times R$ is the total number of QA pairs.

\paragraph{Step 2: Consistency Checking.}  

To overcome the bottleneck of pairwise checking, we design an efficient and scalable prompt-based method named ``You Only Prompt Once'' (\yopo) that reduces inference complexity from quadratic to constant time, i.e., $\mathcal{O}(N^2) \rightarrow \mathcal{O}(1)$. Specifically, \yopo~treats the QA pair matrix as a list of $M$ QA pairs, 
\begin{equation*}
    \mathbb{QA}_{\textup{pair}}=\{\mathcal{QA}_{\textup{pair}}^{(i)}\}_{i=1}^M, \  \mathcal{QA}_{\textup{pair}}^{(i)}=(x, y_{\lfloor\frac{i}{R}\rfloor}^{i-R\lfloor\frac{i}{R}\rfloor}).
\end{equation*}

Then a prompt instruction is devised to check if a specific QA pair, $\mathcal{QA}_{\textup{pair}}^{(i)}$, is semantically equivalent to other QA pairs, $\mathcal{QA}_{\textup{pair}}^{(j)}, j \neq i$, and return the number of QA pairs are semantically equivalent to $\mathcal{QA}_{\textup{pair}}^{(i)}$.  In particular, the prompt instruction guides an LLM to automatically check the pairwise consistency between $\mathcal{QA}_{\textup{pair}}^{(1)}$ and $\{\mathcal{QA}_{\textup{pair}}^{(2)},...,\mathcal{QA}_{\textup{pair}}^{(M)} \}$, then check the consistency between $\mathcal{QA}_{\textup{pair}}^{(2)}$ and $\{\mathcal{QA}_{\textup{pair}}^{(3)},...,\mathcal{QA}_{\textup{pair}}^{(M)}\}$, and finally check the consistency between $\mathcal{QA}_{\textup{pair}}^{(M-1)}$ and $\{\mathcal{QA}_{\textup{pair}}^{(M)}\}$. Instead of executing individual model inferences sequentially, we observe that prominent LLMs can achieve accurate pairwise checking in parallel by prompting once. This is achieved by prompting LLM (see more details in Appendix \ref{sec:prompt_set}) which has been carefully tuned and tested on various cases. 

\paragraph{Step 3: Voting.} 
Leveraging the benefits of \yopo, we collect several QA pairs that are semantically equivalent to a specific QA pair, $\mathcal{QA}_{\textup{pair}}^{(i)}$, denoted by $\mathcal{V}_i$, which is referred to as the number of consistency votes. Then we rank all the votes $\mathbb{V} = \{ \mathcal{V}_i,...,\mathcal{V}_M \}$, return the index of QA pairs with the most and least votes, and finally identify the most consistent response $y^{mc}$ and least consistent response $y^{lc}$. Considering that there might be multiple responses with the most votes, we build a top-ranked set $\mathbb{Y}^{mc} = \{y_1^{mc}, ...,y_K^{mc} \}$ that contains the top $K$ most consistent responses. Similarly, we have the least vote set, $\mathbb{Y}^{lc} = \{y_1^{lc}, ...,y_K^{lc} \}$ that can be used for detecting falseness/incorrectness.

\subsection{\sacsummary: Generative Summary}

The effectiveness of \saccheck~is constrained by the quality of selections from the candidate set $\mathbb{Y}^{mc}$ with the most votes. A simple selection may result in a biased and incomplete response. We hypothesize that merging multiple top-ranked candidate responses can mitigate this constraint. As these top candidates often showcase complementary strengths and weaknesses, it is plausible to generate a superior response by combining their advantages while mitigating their shortcomings. Our objective is to prompt an LLM that takes input $x$ and $K$ top-ranked consistent candidate $ \{ y_1^{mc},...,y_K^{mc}\} \subset \mathbb{Y}^{mc}$ and produces an improved output as the final response. 

To achieve this goal, we present \sacsummary, a prompt-based method for fusing a set of candidate responses conditioned on the input to generate an enhanced consistent output. Specifically, we concatenate the input and $K$ candidates using \sacsummary~prompting (see Appendix \ref{sec:prompt_set}), to produce $\hat{y}$ through a generative summary. More discussion about \sacsummary~is provided in Appendix \ref{sec:summary}. 

\section{Experiments}
In this section, we first evaluate the performance of \saccheck~by comparing it with other baseline methods for consistency checking, then demonstrate the effectiveness of \methodName~on multiple classification and generation tasks. 

\subsection{Evaluation of Consistency Check}
We compare \saccheck~with other baselines for consistency checks, such as BERTscore, n-gram, NLI, and pairwise-prompt that are used by \cite{manakul2023selfcheckgpt} on the hallucination detection task \cite{zhang2023language}.

\paragraph{Dataset and Experiment Setup.} Wikibio hallucination dataset is synthetically generated from Wikibio dataset \cite{lebret2016generating} by manually annotating the factuality of the passage at a sentence level. The dataset consists of 238 annotated passages and 1908 sentences, and each sentence is classified as major inaccurate (39.9\%), minor inaccurate (33.1\%), and accurate (27.0\%). We follow the experimental settings by Manakul \textit{et al.}~\cite{manakul2023selfcheckgpt} to identify the factuality and non-factuality of sentences. Instead of using 20 generated samples, we use 5 samples to reduce computational costs. We perform GPT-3.5 with \yopo~to check the consistency between the target response and candidate responses. 

\paragraph{Baseline Methods}
There are several baseline methods for measuring the consistency between two responses. 
\begin{itemize}[leftmargin=10pt]
    \item {\bf BERTscore} \cite{zhang2019bertscore} is a metric that calculates the similarity between one target response and sampled responses using sentence representation.

\item {\bf n-gram} \cite{brown1992class} aims to train a new language model using generated samples to approximate the blackbox LLM.  
The LLM's token probabilities can thus be approximated using the trained proxy language model. The log probability of the target response is then used as a metric for consistency evaluation. 

\item {\bf Natural Language Inference (NLI)} \cite{kuhn2023semantic} provides entailment score for checking consistency. In our implementation, we use a {DeBERTa-v3-large} model \cite{he2020deberta} that is fine-tuned on the MNLI dataset \cite{williams2017broad} as the entailment model and the normalized probability of entailment classes is used as the final score. 

\item {\bf Pairwise Comparison} \cite{tian2023just} is a prompt-based approach for consistency check. Let $\mathcal{C}(\cdot,\cdot)$ denote an operator for checking semantic equivalence, which takes two QA pairs as input, i.e., $\mathcal{C}((x,y_i),(x,y_j))$. The operator $\mathcal{C}$ returns ``Yes'' if the two QA pairs are semantically equivalent, and ``No'' otherwise. We implement this checking operator using an LLM by leveraging the prompt: \promptgpt{Are the following two QA pairs semantically equivalent? [QA PAIR 1] [QA PAIR 2]}. However, such a vanilla pairwise consistency check is very inefficient as it requires $M\times(M-1)/2$ times model inference for checking the consistency among the QA pair matrix in Eq.~\eqref{eq:matrix}. As a result, this type of approach is difficult to scale as its computational cost grows quadratically with $M$.
\end{itemize}

\paragraph{Results.} 
Table \ref{tab:wikibio} presents the AUC-PR (Precision-Recall) performance of \saccheck~against baseline methods which are typically used for checking consistency. Our results match the observation from prior work \cite{manakul2023selfcheckgpt} where the pairwise-prompt method performs better than the other baselines at the cost of higher computational overheads as the time complexity grows quadratically with the number of responses. The high expenses of multiple LLM API calls raise another concern. In contrast, \saccheck~offers an efficient and cost-effective solution that not only outperforms all the baselines in terms of detection performance but also significantly reduces computation and expense costs.  

\begin{table}[!ht]
\centering
\small
\begin{tabular}{@{}c|ccc@{}}
\toprule
Methods              & \begin{tabular}[c]{@{}c@{}}NonFact \\ (AUC-PR)\end{tabular} & \begin{tabular}[c]{@{}c@{}}Factual \\ (AUC-PR)\end{tabular} & \begin{tabular}[c]{@{}c@{}}Ranking \\ (PCC)\end{tabular} \\ \midrule
BERTScore & 80.23              & 42.07              & 56.22             \\

n-gram   & 83.87              & 55.19              & 62.34             \\
NLI       & 90.23               & 64.90              & 71.78             \\
Pairwise-Prompt    & 91.29              & 65.87              & {\bf 76.64}             \\ \midrule
\saccheck~(\yopo)         & {\bf 91.43}             & {\bf 66.13}              & 76.61             \\ \bottomrule
\end{tabular}
\vspace{-1mm}
  \caption{AUC-PR performance on the hallucination detection task using various consistency check approaches.}
\label{tab:wikibio}
\end{table}

\subsection{Evaluation of \methodName~Framework}
\paragraph{Experiment Setup.} We use three ($N=3$) state-of-the-art blackbox LLMs, GPT-3.5, GPT-4, and PaLM 2 where the evaluations are conducted using Azure OpenAI API (GPT-3.5 and GPT-4) and Google Cloud API (PaLM 2).  When performing model inference, we set the temperature to $0.0$ to generate a single deterministic response and sample four stochastic responses by increasing the temperature to $1.0$.  Therefore, a total of five ($R=5$) output responses are drawn from each of the models such that we have up to 15 candidate QA pairs. We use GPT-3.5 as a proxy to perform \saccheck~with \yopo~to check consistency among all QA pairs and then employ \sacsummary~to generate enhanced output response. 

\begin{table*}[h!]
\centering
\small
\begin{tabular}{l|cc|cc|cc|cc}
\toprule
Methods & \multicolumn{2}{c}{Primality} & \multicolumn{2}{c}{Senator} & \multicolumn{2}{c}{HotpotQA} & \multicolumn{2}{c}{NQ-Open} \\ 

            & Single       & 5 Samples       & Single          & 5 Samples         & Single          & 5 Samples   & Single           & 5 Samples    \\ \midrule
GPT-3.5 (\texttt{gpt-3.5-turbo})      & 37.2        & 44.4          & 13.7           & {\bf 31.0}            & 48.8           & 48.0      & 74.1            & 69.6       \\
GPT-4 (\texttt{gpt-4})        & 27.2        & 41.8          & 2.1           & 3.8            & 68.0    & 70.8      & 83.2   & 81.6       \\
PaLM 2 (\texttt{chat-bison})        & 15.4        & 24.2          & 3.6           & 16.9            & 52.8           & 65.6      & 58.0            & 72.8       \\ \midrule
\methodName~(proposed method)   &      /        & {\bf 62.8}          &       /          & 24.0            &    /             & {\bf 77.2}      &            /      & {\bf 85.7}       \\ 
\bottomrule
\end{tabular}

  \caption{Empirical comparison of LLM responses on truthfulness accuracy (\%) across four benchmark datasets.} 
\label{tab:main}
\end{table*}

\paragraph{Evaluation Metrics.}
(1) \textit{Truthfulness Accuracy}: To evaluate the truthfulness of the generated responses, we measure the rate of the output responses being semantically consistent (as defined in Eq.\eqref{eq:score}) with the ground truth. 
(2) \textit{Falseness Accuracy}: Conversely, we also measure the rate of the least consistent response produced by \saccheck~being semantically inequivalent to the ground truth, as a metric for detecting non-factuality and hallucination.

\paragraph{Datasets.}
We evaluate our \methodName~approach on two categories of QA tasks, namely, classification QA and generation QA, with each category containing two datasets. Following prior work~\cite{zhang2023language}, we use the following two binary classification datasets: prime number ({\bf Primality}) and senator search ({\bf Senator}) \cite{zhang2023language} for evaluation. For the generation tasks, we take questions from two open-domain QA datasets, {\bf HotpotQA} \cite{yang2018hotpotqa} and {\bf NQ-open} \cite{lee2019latent}, and generate answers using LLMs. Then we manually annotate the factuality of the answers following previous work~\cite{li2023halueval}. The details of both datasets are provided in Appendix \ref{sec:dataset_details}. 

\begin{figure}[h!]
\centering
    % \vspace{-3pt}
    \includegraphics[width=0.23\textwidth]{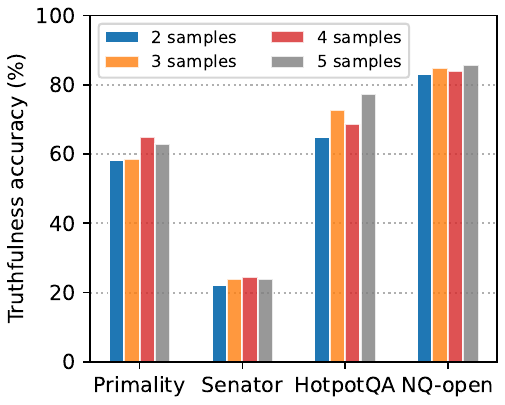} 
    \includegraphics[width=0.23\textwidth]{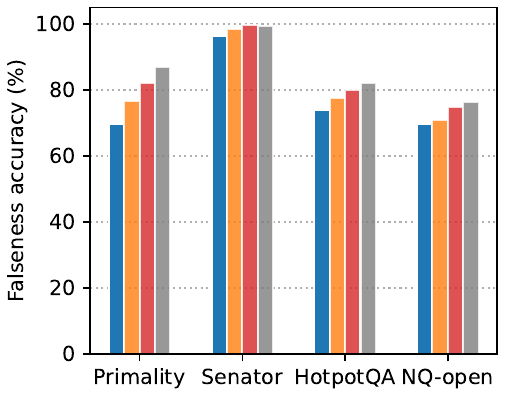} 
    \caption{The effect of sample size on the truthfulness accuracy (left) with the most consistency votes and falseness accuracy (right) with the least consistency votes.}
    \label{fig:effect}
    % \vspace{-2mm}
    % \vspace{-10pt}
\end{figure}

\paragraph{Main Results.}
In Table \ref{tab:main}, we present the overall performance of \methodName~as well as other methods with the single model on four benchmark datasets. For each model, i.e., GPT-3.5, GPT-4, and PaLM 2, we use the single deterministic responses (with the temperature set to $0.0$) as the baseline. For comparison, we also generate four additional samples (with the temperature set to $1.0$) for each LLM and select the response with the most consistent votes as the final response. We observed that the result of multiple samples consistently outperforms the single response, specifically in Primality and Senator datasets. Our proposed \methodName~approach significantly outperforms any single model with sampled responses by a large margin in Primality (+18.8), HotpotQA (+6.4), and NQ-open (+4.1) datasets. It is interesting to observe that \methodName~is suppressed by GPT-3.5 but performs better than GPT-4 and PaLM 2 in the senator task. This is because GPT-4 and PaLM 2 perform inconsistent responses, but occupy the majority of voting, hurting the final performance of LLM ensembles. 

\paragraph{Most Consistent vs Least Consistent.} The core capability of \methodName~is to ensemble these LLMs' outputs to generate consistently reliable responses for each input. Beyond that, \methodName~also allows us to identify the most inconsistent responses with the least votes from candidate models. It is thus worthy to evaluate if these inconsistent responses are truly non-factual or hallucinated, i.e., falseness in general. Table \ref{tab:most_inconsistent} shows the results of the falseness accuracy based on 5 sampled responses. Our ensemble method performs best in detecting falseness against other single models on all datasets. 
\begin{table}[h!]

\small
\centering
\begin{tabular}{@{}c|cccc@{}}
\toprule
Methods & Primality & Senator & HotpotQA & NQ-open \\ \midrule
GPT-3.5       & 78.0     & 83.5          & 67.2    & 44.6      \\
GPT-4         & 62.4     & 97.2          & 36.8    & 25.2      \\
PaLM 2         & 60.2     & 92.5          & 63.2    & 60.4      \\  \midrule
\methodName      & {\bf 87.0}     & {\bf 99.4}          & {\bf 82.0}    & {\bf 76.3}     \\ \bottomrule
\end{tabular}

  \caption{Empirical comparison of different LLMs' responses on falseness accuracy (\%) using the least consistent votes.}
\label{tab:most_inconsistent}
\end{table}

\begin{figure}[h!]
\centering
    \includegraphics[width=0.47\textwidth]{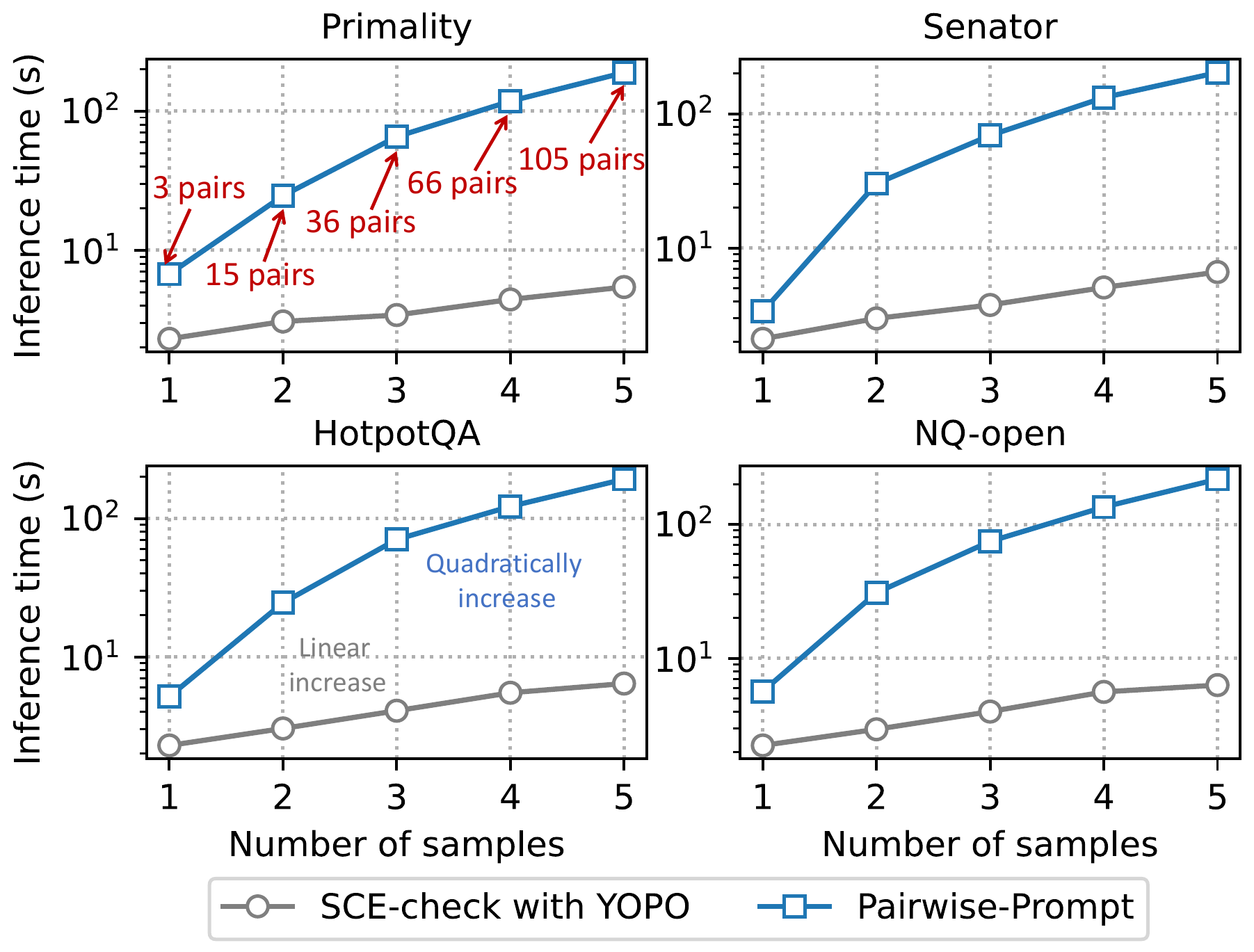} 
    \caption{Scalability analysis of consistency check.}
    \label{fig:scalable}
    \vspace{-5mm}
\end{figure}

\paragraph{Effect of the Number of Responses.} The performance of sampling-based methods is expected to improve as the sample size increases. The mechanism of consistent voting also relies on the number of generated responses. Fig.~\ref{fig:effect} (left) presents the truthfulness accuracy under different numbers of sampled responses ranging from two samples to five samples. Although using more samples slightly improves the performance, on most datasets we observe such improvement is marginal compared to the increase in computational costs. This suggests that in practice, we could use a sample size of as small as two to achieve reasonably good performance. On the contrary, the falseness results in Fig. \ref{fig:effect} (right), show a clear trend, i.e., the more samples, the better performance. This is because more samples increase the probability of generating inconsistent responses in general.

\begin{figure*}[h!]
\centering
    \includegraphics[width=0.24\textwidth]{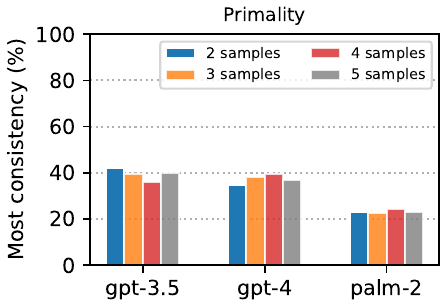}    \includegraphics[width=0.24\textwidth]{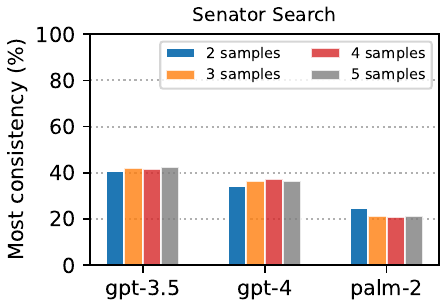}
    \includegraphics[width=0.24\textwidth]{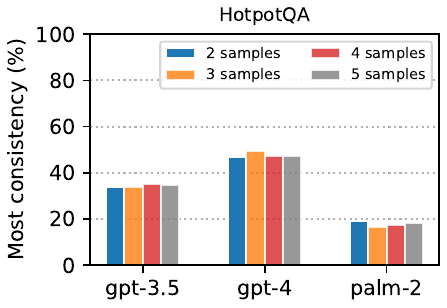}
    \includegraphics[width=0.24\textwidth]{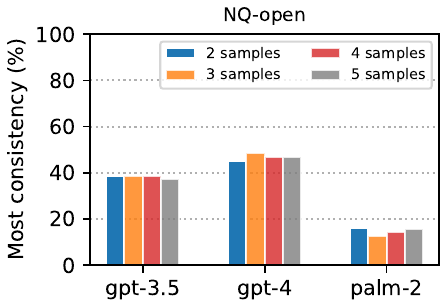}
    \vspace{-2.5mm}
    \caption{Effect of sample size on the proportion of the most consistent votes in model ensembles across four datasets.}
    \label{fig:consistency_sample}
\end{figure*}

\begin{figure*}[h!]
\centering
    \vspace{-2mm}
    \includegraphics[width=0.24\textwidth]{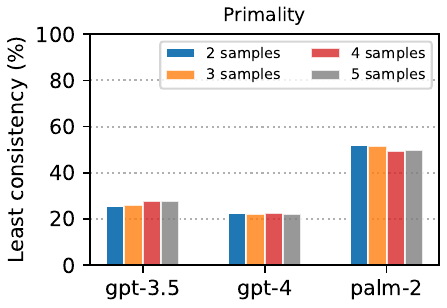}    \includegraphics[width=0.24\textwidth]{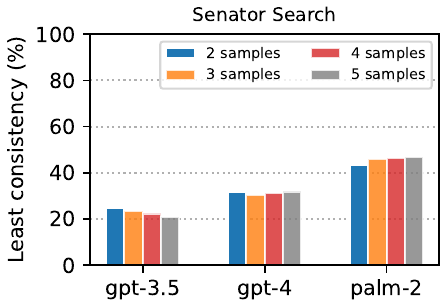}
    \includegraphics[width=0.24\textwidth]{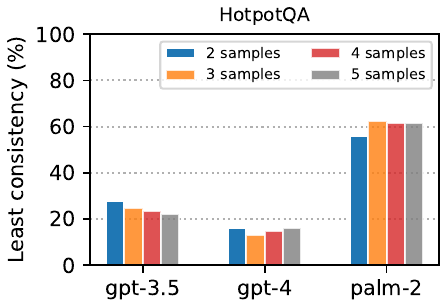}
    \includegraphics[width=0.24\textwidth]{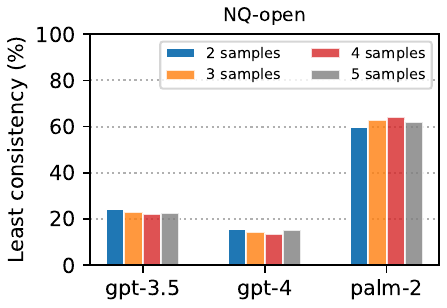}
    \vspace{-2.5mm}
    \caption{Effect of sample size on the proportion of the least consistent votes in model ensembles across four datasets.}
    \label{fig:inconsistency_sample}
    \vspace{-3mm}
\end{figure*}

\paragraph{Scalability Analysis.} While more samples potentially increase the performance, this certainly leads to a very high computational cost for consistency checking among these responses. Conventional pairwise consistency check, either using prompt or similarity metrics, e.g., BERTscore, is infeasible to scale due to high complexity $\mathcal{O}(N^2)$ as the resulting inference time increases quadratically. Our \methodName~overcomes these challenges by checking all consistency and inconsistency by prompting the model once. As shown in Fig. \ref{fig:scalable}, \saccheck~shows linear scaling of inference time due to increased token length (as the sample size grows). \saccheck~with \yopo~achieves significant time reduction by two orders of magnitude compared with the pairwise-prompt methods. The scalability analysis of four datasets demonstrates a consistent trend of linear scaling of our scalable consistency check, which can be applied to wide tasks beyond what's included in our study.

\paragraph{Which Model is More Consistent?} It is crucial to understand the diverse strengths and weaknesses of LLMs when performing an ensembling method. For all responses with the most consistent votes, we present a study to analyze the proportion of votes from each model, as shown in Table \ref{tab:consistent_vote}. Note that GPT-3.5 shows a higher proportion than GPT-4 (+2.7, +6.2) and PaLM 2 (+16.7, +21.3) in Primality and Senator datasets while GPT-4 suppresses the other two models in HotpotQA and NQ-open datasets. On average, GPT-4 performs slightly better than GPT-3.5 (+2.8) but both GPT-based models outperform PaLM 2 in these tasks. 

\begin{table}[h!]
\resizebox{\linewidth}{!}{
\begin{tabular}{@{}c|cccc|c@{}}
\toprule
Methods & Primality & Senator & HotpotQA & NQ-open& Avg \\ \midrule
GPT-3.5     & {\bf 39.8}    & {\bf 42.5}         & 34.8   & 37.5     & 38.6  \\
GPT-4       & 37.1    & 36.3         & {\bf 47.1}   & {\bf 46.8}     & {\bf 41.8}  \\
PaLM 2       & 23.1    & 21.2         & 18.1   & 15.7     & 19.5  \\ \bottomrule
\end{tabular}
}
  \caption{Proportion (\%) of the most consistent votes.}
\label{tab:consistent_vote}
\vspace{-3mm}
\end{table}

\begin{table}[h!]
\resizebox{\linewidth}{!}{
\begin{tabular}{@{}c|cccc|c@{}}
\toprule
Methods & Primality & Senator & HotpotQA & NQ-open & Avg \\ \midrule
GPT-3.5       & 27.9    & 21.1         & 22.3   & 22.5     & 23.4  \\
GPT-4         & 22.2    & 31.9         & 16.0   & 15.3     & 21.4  \\
PaLM 2         & {\bf 49.9}    & {\bf 47.0}         & {\bf 61.7}   & {\bf 62.2}     & {\bf 55.2}  \\ \bottomrule
\end{tabular}
}
  \caption{Proportion (\%) of the least consistent votes.}
\label{tab:inconsistent_vote}
\end{table}

Table \ref{tab:inconsistent_vote} reports the results of least consistent votes, where PaLM 2 presents a much higher proportion than GPT-3.5 and GPT-4, which reflects that the responses from PaLM 2 are often inconsistent with the other two models, specifically on HotpotQA and NQ-open datasets. In contrast, GPT-4 and GPT-3.5 show a similarly low proportion of the least inconsistent votes on average.  Fig. \ref{fig:consistency_sample} and \ref{fig:inconsistency_sample} present an ablation study to analyze the effect of sample size on the most and least consistent votes respectively. We observe the proportion of each model is relatively stable, where the performance is insensitive to the sample size. From this analysis, we note that GPT-4 performs better consistent and reliable responses while PaLM 2 shows inconsistency against the other models. 

\section{Related Work}
\paragraph{Consistency Evaluation of LLMs.} 
An essential characteristic of logically reliable intelligent systems is self-consistency, which entails that no two statements provided by the system contradict each other. Self-consistency is defined by \citet{elazar2021measuring} as the invariance of an LLM's responses across various types of semantics-preserving transformations. This definition is enriched by multiple other consistency categories proposed by \citet{jang2022becel}. \citet{wang2022self} demonstrates that self-consistency can significantly enhance the chain of thought reasoning in LLMs. Without self-consistency, it becomes challenging to regard LLMs as a reliable or trustworthy system \cite{li2025cv, cui2024dcr}. Recent studies employ self-consistency to detect hallucinations based on pre-trained LLMs \cite{manakul2023selfcheckgpt} and instruction-tuned LLMs \cite{mundler2023self}. Although these methods exhibit promising accuracy on several benchmark tasks, the efficiency of pairwise checking/ranking has been a major concern due to $\mathcal{O}(N^2)$ model calls for getting the full matrix \cite{jiang2023llm,zheng2023judging,tian2023just}. These less efficient solutions severely limit the potential for performing consistency evaluations. In contrast, Our \saccheck~approach successfully breaks through this barrier by significantly reducing the time complexity while achieving competitive performance.  

\paragraph{Ensemble Learning in LLMs.} Ensemble methods reliably improve performance in a number of contexts by leveraging multiple models \cite{lakshminarayanan2017simple,ganaie2022ensemble}. Typically, ensemble learning is performed either by considering model weights or by combing diverse outputs \cite{liang2022camero}. Mix-of-Experts (MoE) is often employed to combine the predictions of multiple specialized sub-models to improve overall performance and has been widely applied in various LLM domains \cite{du2022glam, artetxe2021efficient,shen2023flan}. Existing studies have shown that ensembling prompts \cite{wang2022rationale} is more efficient than ensembling fine-tuned models \cite{lester2021power} and some recent methods adopt prompt boosting for achieving better performance \cite{hou2023promptboosting, pitis2023boosted}. Such method requires drafting multiple prompts, which lead to additional overhead considering the sensitivity of LLMs on prompts \cite{cui2024phaseevo, sinha2024sos, cui2025aposurvay}. Unlike the MoE or prompt ensemble methods, our proposed method referred to as \methodName, dynamically ensemble outputs drawn from multiple prominent LLMs to generate consistently better responses that harness the complementary potentials, leading to improved reliability and robustness. A concurrent work \cite{jiang2023llm}, seeks a similar goal by leveraging multiple open-source LLMs but its method is a much less efficient solution and strongly relies on additional pre-trained models. In contrast, our proposed method offers a scalable and flexible solution with prompt only without requiring any models. 

\section{Conclusion}
We design an efficient and scalable ensemble framework for generating reliable responses by leveraging the complementarity of prominent blackbox LLMs. By combining two key components \saccheck~and \sacsummary, our approach consistently outperforms existing baselines in terms of effectiveness and efficiency as illustrated by our extensive experiments on five datasets.  

We note that \saccheck~with \yopo~achieves comparable performance as the pairwise check via prompt while significantly reducing the computational cost by two orders of magnitude, which would benefit researchers and practitioners in real-world LLM deployment and applications. 

\section*{Limitations}

Although our approach shows promising performance, there are several \textbf{limitations}: (1) \textit{Model Selection}. It is critical to identify an appropriate set of LLM candidates. Choosing an LLM that doesn't fit the task may pollute the consistent votes and subsequently hurt final performance; (2) \textit{Model Diversity}. Our approach benefits from employing a diverse set of LLMs for ensembling. Although our current implementation only considered three LLMs, our framework naturally generalizes to a larger number of LLMs to further improve the performance; and (3) \textit{Human Evaluation}. Verifying the truthfulness of the responses is a non-trivial task and is crucial for evaluation. Due to practical considerations, in our evaluation, we employed GPT-4 as a proxy for human evaluation, which may not always be accurate. Future work may explore a hybrid strategy to incorporate both humans and LLMs to conduct a more comprehensive evaluation.

\section*{Ethics Statement}
This work improves the reliability of LLM generation via scalable consistency ensembles of different LLMs. It may help to mitigate the generation of hallucinations and the use of mistaken or misleading content from LLMs, to encourage a safer use of these models.

\bibliography{anthology,custom}

\begin{thebibliography}{48}
\expandafter\ifx\csname natexlab\endcsname\relax\def\natexlab#1{#1}\fi

\bibitem[{Anil et~al.(2023)Anil, Dai, Firat, Johnson, Lepikhin, Passos, Shakeri, Taropa, Bailey, Chen et~al.}]{anil2023palm}
Rohan Anil, Andrew~M Dai, Orhan Firat, Melvin Johnson, Dmitry Lepikhin, Alexandre Passos, Siamak Shakeri, Emanuel Taropa, Paige Bailey, Zhifeng Chen, et~al. 2023.
\newblock Palm 2 technical report.
\newblock \emph{arXiv preprint arXiv:2305.10403}.

\bibitem[{Artetxe et~al.(2021)Artetxe, Bhosale, Goyal, Mihaylov, Ott, Shleifer, Lin, Du, Iyer, Pasunuru et~al.}]{artetxe2021efficient}
Mikel Artetxe, Shruti Bhosale, Naman Goyal, Todor Mihaylov, Myle Ott, Sam Shleifer, Xi~Victoria Lin, Jingfei Du, Srinivasan Iyer, Ramakanth Pasunuru, et~al. 2021.
\newblock Efficient large scale language modeling with mixtures of experts.
\newblock \emph{arXiv preprint arXiv:2112.10684}.

\bibitem[{Biderman et~al.(2023)Biderman, Schoelkopf, Anthony, Bradley, O’Brien, Hallahan, Khan, Purohit, Prashanth, Raff et~al.}]{biderman2023pythia}
Stella Biderman, Hailey Schoelkopf, Quentin~Gregory Anthony, Herbie Bradley, Kyle O’Brien, Eric Hallahan, Mohammad~Aflah Khan, Shivanshu Purohit, USVSN~Sai Prashanth, Edward Raff, et~al. 2023.
\newblock Pythia: A suite for analyzing large language models across training and scaling.
\newblock In \emph{International Conference on Machine Learning}, pages 2397--2430. PMLR.

\bibitem[{Brown et~al.(1992)Brown, Della~Pietra, Desouza, Lai, and Mercer}]{brown1992class}
Peter~F Brown, Vincent~J Della~Pietra, Peter~V Desouza, Jennifer~C Lai, and Robert~L Mercer. 1992.
\newblock Class-based n-gram models of natural language.
\newblock \emph{Computational linguistics}, 18(4):467--480.

\bibitem[{Chiang et~al.(2023)Chiang, Li, Lin, Sheng, Wu, Zhang, Zheng, Zhuang, Zhuang, Gonzalez et~al.}]{chiang2023vicuna}
Wei-Lin Chiang, Zhuohan Li, Zi~Lin, Ying Sheng, Zhanghao Wu, Hao Zhang, Lianmin Zheng, Siyuan Zhuang, Yonghao Zhuang, Joseph~E Gonzalez, et~al. 2023.
\newblock Vicuna: An open-source chatbot impressing gpt-4 with 90\%* chatgpt quality.
\newblock \emph{See https://vicuna. lmsys. org (accessed 14 April 2023)}.

\bibitem[{Chung et~al.(2022)Chung, Hou, Longpre, Zoph, Tay, Fedus, Li, Wang, Dehghani, Brahma et~al.}]{chung2022scaling}
Hyung~Won Chung, Le~Hou, Shayne Longpre, Barret Zoph, Yi~Tay, William Fedus, Eric Li, Xuezhi Wang, Mostafa Dehghani, Siddhartha Brahma, et~al. 2022.
\newblock Scaling instruction-finetuned language models.
\newblock \emph{arXiv preprint arXiv:2210.11416}.

\bibitem[{Conover et~al.(2023)Conover, Hayes, Mathur, Xie, Wan, Shah, Ghodsi, Wendell, Zaharia, and Xin}]{DatabricksBlog2023DollyV2}
Mike Conover, Matt Hayes, Ankit Mathur, Jianwei Xie, Jun Wan, Sam Shah, Ali Ghodsi, Patrick Wendell, Matei Zaharia, and Reynold Xin. 2023.
\newblock \href {https://www.databricks.com/blog/2023/04/12/dolly-first-open-commercially-viable-instruction-tuned-llm} {Free dolly: Introducing the world's first truly open instruction-tuned llm}.

\bibitem[{Cui et~al.(2024{\natexlab{a}})Cui, Li, Lopez, Das, Malin, Kumar, and Zhang}]{cui2024dcr}
Wendi Cui, Zhuohang Li, Damien Lopez, Kamalika Das, Bradley~A. Malin, Sricharan Kumar, and Jiaxin Zhang. 2024{\natexlab{a}}.
\newblock \href {https://doi.org/10.18653/v1/2024.emnlp-industry.25} {Divide-conquer-reasoning for consistency evaluation and automatic improvement of large language models}.
\newblock In \emph{Proceedings of the 2024 Conference on Empirical Methods in Natural Language Processing: Industry Track}, pages 334--361, Miami, Florida, US. Association for Computational Linguistics.

\bibitem[{Cui et~al.(2024{\natexlab{b}})Cui, Zhang, Li, Sun, Lopez, Das, Malin, and Kumar}]{cui2024phaseevo}
Wendi Cui, Jiaxin Zhang, Zhuohang Li, Hao Sun, Damien Lopez, Kamalika Das, Bradley Malin, and Sricharan Kumar. 2024{\natexlab{b}}.
\newblock Phaseevo: Towards unified in-context prompt optimization for large language models.
\newblock \emph{arXiv preprint arXiv:2402.11347}.

\bibitem[{Cui et~al.(2025)Cui, Zhang, Li, Sun, Lopez, Das, Malin, and Kumar}]{cui2025aposurvay}
Wendi Cui, Jiaxin Zhang, Zhuohang Li, Hao Sun, Damien Lopez, Kamalika Das, Bradley~A. Malin, and Sricharan Kumar. 2025.
\newblock Automatic prompt optimization via heuristic search: A survey.
\newblock \emph{arXiv preprint arXiv:2502.18746}.

\bibitem[{Du et~al.(2022)Du, Huang, Dai, Tong, Lepikhin, Xu, Krikun, Zhou, Yu, Firat et~al.}]{du2022glam}
Nan Du, Yanping Huang, Andrew~M Dai, Simon Tong, Dmitry Lepikhin, Yuanzhong Xu, Maxim Krikun, Yanqi Zhou, Adams~Wei Yu, Orhan Firat, et~al. 2022.
\newblock Glam: Efficient scaling of language models with mixture-of-experts.
\newblock In \emph{International Conference on Machine Learning}, pages 5547--5569. PMLR.

\bibitem[{Elazar et~al.(2021)Elazar, Kassner, Ravfogel, Ravichander, Hovy, Sch{\"u}tze, and Goldberg}]{elazar2021measuring}
Yanai Elazar, Nora Kassner, Shauli Ravfogel, Abhilasha Ravichander, Eduard Hovy, Hinrich Sch{\"u}tze, and Yoav Goldberg. 2021.
\newblock Measuring and improving consistency in pretrained language models.
\newblock \emph{Transactions of the Association for Computational Linguistics}, 9:1012--1031.

\bibitem[{Ganaie et~al.(2022)Ganaie, Hu, Malik, Tanveer, and Suganthan}]{ganaie2022ensemble}
Mudasir~A Ganaie, Minghui Hu, AK~Malik, M~Tanveer, and PN~Suganthan. 2022.
\newblock Ensemble deep learning: A review.
\newblock \emph{Engineering Applications of Artificial Intelligence}, 115:105151.

\bibitem[{He et~al.(2020)He, Liu, Gao, and Chen}]{he2020deberta}
Pengcheng He, Xiaodong Liu, Jianfeng Gao, and Weizhu Chen. 2020.
\newblock Deberta: Decoding-enhanced bert with disentangled attention.
\newblock \emph{arXiv preprint arXiv:2006.03654}.

\bibitem[{Hou et~al.(2023)Hou, O’connor, Andreas, Chang, and Zhang}]{hou2023promptboosting}
Bairu Hou, Joe O’connor, Jacob Andreas, Shiyu Chang, and Yang Zhang. 2023.
\newblock Promptboosting: Black-box text classification with ten forward passes.
\newblock In \emph{International Conference on Machine Learning}, pages 13309--13324. PMLR.

\bibitem[{Huang et~al.(2022)Huang, Gu, Hou, Wu, Wang, Yu, and Han}]{huang2022large}
Jiaxin Huang, Shixiang~Shane Gu, Le~Hou, Yuexin Wu, Xuezhi Wang, Hongkun Yu, and Jiawei Han. 2022.
\newblock Large language models can self-improve.
\newblock \emph{arXiv preprint arXiv:2210.11610}.

\bibitem[{Jang et~al.(2022)Jang, Kwon, and Lukasiewicz}]{jang2022becel}
Myeongjun Jang, Deuk~Sin Kwon, and Thomas Lukasiewicz. 2022.
\newblock Becel: Benchmark for consistency evaluation of language models.
\newblock In \emph{Proceedings of the 29th International Conference on Computational Linguistics}, pages 3680--3696.

\bibitem[{Jiang et~al.(2023)Jiang, Ren, and Lin}]{jiang2023llm}
Dongfu Jiang, Xiang Ren, and Bill~Yuchen Lin. 2023.
\newblock Llm-blender: Ensembling large language models with pairwise ranking and generative fusion.
\newblock \emph{arXiv preprint arXiv:2306.02561}.

\bibitem[{K{\"o}pf et~al.(2023)K{\"o}pf, Kilcher, von R{\"u}tte, Anagnostidis, Tam, Stevens, Barhoum, Duc, Stanley, Nagyfi et~al.}]{kopf2023openassistant}
Andreas K{\"o}pf, Yannic Kilcher, Dimitri von R{\"u}tte, Sotiris Anagnostidis, Zhi-Rui Tam, Keith Stevens, Abdullah Barhoum, Nguyen~Minh Duc, Oliver Stanley, Rich{\'a}rd Nagyfi, et~al. 2023.
\newblock Openassistant conversations--democratizing large language model alignment.
\newblock \emph{arXiv preprint arXiv:2304.07327}.

\bibitem[{Kuhn et~al.(2023)Kuhn, Gal, and Farquhar}]{kuhn2023semantic}
Lorenz Kuhn, Yarin Gal, and Sebastian Farquhar. 2023.
\newblock Semantic uncertainty: Linguistic invariances for uncertainty estimation in natural language generation.
\newblock \emph{arXiv preprint arXiv:2302.09664}.

\bibitem[{Kwiatkowski et~al.(2019)Kwiatkowski, Palomaki, Redfield, Collins, Parikh, Alberti, Epstein, Polosukhin, Devlin, Lee et~al.}]{kwiatkowski2019natural}
Tom Kwiatkowski, Jennimaria Palomaki, Olivia Redfield, Michael Collins, Ankur Parikh, Chris Alberti, Danielle Epstein, Illia Polosukhin, Jacob Devlin, Kenton Lee, et~al. 2019.
\newblock Natural questions: a benchmark for question answering research.
\newblock \emph{Transactions of the Association for Computational Linguistics}, 7:453--466.

\bibitem[{Lakshminarayanan et~al.(2017)Lakshminarayanan, Pritzel, and Blundell}]{lakshminarayanan2017simple}
Balaji Lakshminarayanan, Alexander Pritzel, and Charles Blundell. 2017.
\newblock Simple and scalable predictive uncertainty estimation using deep ensembles.
\newblock \emph{Advances in neural information processing systems}, 30.

\bibitem[{Lebret et~al.(2016)Lebret, Grangier, and Auli}]{lebret2016generating}
R{\'e}mi Lebret, David Grangier, and Michael Auli. 2016.
\newblock Generating text from structured data with application to the biography domain.
\newblock \emph{ArXiv e-prints, March}.

\bibitem[{Lee et~al.(2019)Lee, Chang, and Toutanova}]{lee2019latent}
Kenton Lee, Ming-Wei Chang, and Kristina Toutanova. 2019.
\newblock Latent retrieval for weakly supervised open domain question answering.
\newblock In \emph{Proceedings of the 57th Annual Meeting of the Association for Computational Linguistics}, pages 6086--6096.

\bibitem[{Lester et~al.(2021)Lester, Al-Rfou, and Constant}]{lester2021power}
Brian Lester, Rami Al-Rfou, and Noah Constant. 2021.
\newblock The power of scale for parameter-efficient prompt tuning.
\newblock \emph{arXiv preprint arXiv:2104.08691}.

\bibitem[{Li et~al.(2023)Li, Cheng, Zhao, Nie, and Wen}]{li2023halueval}
Junyi Li, Xiaoxue Cheng, Wayne~Xin Zhao, Jian-Yun Nie, and Ji-Rong Wen. 2023.
\newblock Halueval: A large-scale hallucination evaluation benchmark for large language models.
\newblock \emph{arXiv e-prints}, pages arXiv--2305.

\bibitem[{Li et~al.(2025)Li, Yan, Jackson, Cui, Li, Zhang, and Malin}]{li2025cv}
Zhuohang Li, Chao Yan, Nicholas~J. Jackson, Wendi Cui, Bo~Li, Jiaxin Zhang, and Bradley~A. Malin. 2025.
\newblock Towards statistical factuality guarantee for large vision-language models.
\newblock \emph{arXiv preprint arXiv:2502.20560}.

\bibitem[{Liang et~al.(2022)Liang, He, Shen, Chen, and Zhao}]{liang2022camero}
Chen Liang, Pengcheng He, Yelong Shen, Weizhu Chen, and Tuo Zhao. 2022.
\newblock Camero: Consistency regularized ensemble of perturbed language models with weight sharing.
\newblock \emph{arXiv preprint arXiv:2204.06625}.

\bibitem[{Manakul et~al.(2023)Manakul, Liusie, and Gales}]{manakul2023selfcheckgpt}
Potsawee Manakul, Adian Liusie, and Mark~JF Gales. 2023.
\newblock Selfcheckgpt: Zero-resource black-box hallucination detection for generative large language models.
\newblock \emph{arXiv preprint arXiv:2303.08896}.

\bibitem[{M{\"u}ndler et~al.(2023)M{\"u}ndler, He, Jenko, and Vechev}]{mundler2023self}
Niels M{\"u}ndler, Jingxuan He, Slobodan Jenko, and Martin Vechev. 2023.
\newblock Self-contradictory hallucinations of large language models: Evaluation, detection and mitigation.
\newblock \emph{arXiv preprint arXiv:2305.15852}.

\bibitem[{OpenAI(2023)}]{OpenAI2023GPT4TR}
OpenAI. 2023.
\newblock \href {https://api.semanticscholar.org/CorpusID:257532815} {Gpt-4 technical report}.
\newblock \emph{ArXiv}, abs/2303.08774.

\bibitem[{Penedo et~al.(2023)Penedo, Malartic, Hesslow, Cojocaru, Cappelli, Alobeidli, Pannier, Almazrouei, and Launay}]{refinedweb}
Guilherme Penedo, Quentin Malartic, Daniel Hesslow, Ruxandra Cojocaru, Alessandro Cappelli, Hamza Alobeidli, Baptiste Pannier, Ebtesam Almazrouei, and Julien Launay. 2023.
\newblock \href {http://arxiv.org/abs/2306.01116} {The {R}efined{W}eb dataset for {F}alcon {LLM}: outperforming curated corpora with web data, and web data only}.
\newblock \emph{arXiv preprint arXiv:2306.01116}.

\bibitem[{Pitis et~al.(2023)Pitis, Zhang, Wang, and Ba}]{pitis2023boosted}
Silviu Pitis, Michael~R Zhang, Andrew Wang, and Jimmy Ba. 2023.
\newblock Boosted prompt ensembles for large language models.
\newblock \emph{arXiv preprint arXiv:2304.05970}.

\bibitem[{Shen et~al.(2023)Shen, Hou, Zhou, Du, Longpre, Wei, Chung, Zoph, Fedus, Chen et~al.}]{shen2023flan}
Sheng Shen, Le~Hou, Yanqi Zhou, Nan Du, Shayne Longpre, Jason Wei, Hyung~Won Chung, Barret Zoph, William Fedus, Xinyun Chen, et~al. 2023.
\newblock Flan-moe: Scaling instruction-finetuned language models with sparse mixture of experts.
\newblock \emph{arXiv preprint arXiv:2305.14705}.

\bibitem[{Sinha et~al.(2024)Sinha, Cui, Das, and Zhang}]{sinha2024sos}
Ankita Sinha, Wendi Cui, Kamalika Das, and Jiaxin Zhang. 2024.
\newblock \href {https://doi.org/10.18653/v1/2024.emnlp-industry.76} {Survival of the safest: Towards secure prompt optimization through interleaved multi-objective evolution}.
\newblock In \emph{Proceedings of the 2024 Conference on Empirical Methods in Natural Language Processing: Industry Track}, pages 1016--1027, Miami, Florida, US. Association for Computational Linguistics.

\bibitem[{Taori et~al.(2023)Taori, Gulrajani, Zhang, Dubois, Li, Guestrin, Liang, and Hashimoto}]{alpaca}
Rohan Taori, Ishaan Gulrajani, Tianyi Zhang, Yann Dubois, Xuechen Li, Carlos Guestrin, Percy Liang, and Tatsunori~B. Hashimoto. 2023.
\newblock Stanford alpaca: An instruction-following llama model.
\newblock \url{https://github.com/tatsu-lab/stanford_alpaca}.

\bibitem[{Tian et~al.(2023)Tian, Mitchell, Zhou, Sharma, Rafailov, Yao, Finn, and Manning}]{tian2023just}
Katherine Tian, Eric Mitchell, Allan Zhou, Archit Sharma, Rafael Rafailov, Huaxiu Yao, Chelsea Finn, and Christopher~D Manning. 2023.
\newblock Just ask for calibration: Strategies for eliciting calibrated confidence scores from language models fine-tuned with human feedback.
\newblock \emph{arXiv preprint arXiv:2305.14975}.

\bibitem[{Touvron et~al.(2023)Touvron, Martin, Stone, Albert, Almahairi, Babaei, Bashlykov, Batra, Bhargava, Bhosale et~al.}]{touvron2023llama}
Hugo Touvron, Louis Martin, Kevin Stone, Peter Albert, Amjad Almahairi, Yasmine Babaei, Nikolay Bashlykov, Soumya Batra, Prajjwal Bhargava, Shruti Bhosale, et~al. 2023.
\newblock Llama 2: Open foundation and fine-tuned chat models.
\newblock \emph{arXiv preprint arXiv:2307.09288}.

\bibitem[{Wang et~al.(2022{\natexlab{a}})Wang, Wei, Schuurmans, Le, Chi, and Zhou}]{wang2022rationale}
Xuezhi Wang, Jason Wei, Dale Schuurmans, Quoc Le, Ed~Chi, and Denny Zhou. 2022{\natexlab{a}}.
\newblock Rationale-augmented ensembles in language models.
\newblock \emph{arXiv preprint arXiv:2207.00747}.

\bibitem[{Wang et~al.(2022{\natexlab{b}})Wang, Wei, Schuurmans, Le, Chi, and Zhou}]{wang2022self}
Xuezhi Wang, Jason Wei, Dale Schuurmans, Quoc Le, Ed~Chi, and Denny Zhou. 2022{\natexlab{b}}.
\newblock Self-consistency improves chain of thought reasoning in language models.
\newblock \emph{arXiv preprint arXiv:2203.11171}.

\bibitem[{Williams et~al.(2017)Williams, Nangia, and Bowman}]{williams2017broad}
Adina Williams, Nikita Nangia, and Samuel~R Bowman. 2017.
\newblock A broad-coverage challenge corpus for sentence understanding through inference.
\newblock \emph{arXiv preprint arXiv:1704.05426}.

\bibitem[{Yang et~al.(2018)Yang, Qi, Zhang, Bengio, Cohen, Salakhutdinov, and Manning}]{yang2018hotpotqa}
Zhilin Yang, Peng Qi, Saizheng Zhang, Yoshua Bengio, William Cohen, Ruslan Salakhutdinov, and Christopher~D Manning. 2018.
\newblock Hotpotqa: A dataset for diverse, explainable multi-hop question answering.
\newblock In \emph{Proceedings of the 2018 Conference on Empirical Methods in Natural Language Processing}, pages 2369--2380.

\bibitem[{Zhang et~al.(2024)Zhang, Cui, Huang, Das, and Kumar}]{ski2024zhang}
Jiaxin Zhang, Wendi Cui, Yiran Huang, Kamalika Das, and Sricharan Kumar. 2024.
\newblock Synthetic knowledge ingestion: Towards knowledge refinement and injection for enhancing large language models.
\newblock \emph{arXiv preprint arXiv:2410.09629}.

\bibitem[{Zhang et~al.(2023{\natexlab{a}})Zhang, Li, Das, Malin, and Kumar}]{zhang2023sac}
Jiaxin Zhang, Zhuohang Li, Kamalika Das, Bradley~A Malin, and Sricharan Kumar. 2023{\natexlab{a}}.
\newblock Sac$^3$: Reliable hallucination detection in black-box language models via semantic-aware cross-check consistency.
\newblock \emph{arXiv preprint arXiv:2311.01740}.

\bibitem[{Zhang et~al.(2023{\natexlab{b}})Zhang, Press, Merrill, Liu, and Smith}]{zhang2023language}
Muru Zhang, Ofir Press, William Merrill, Alisa Liu, and Noah~A Smith. 2023{\natexlab{b}}.
\newblock How language model hallucinations can snowball.
\newblock \emph{arXiv preprint arXiv:2305.13534}.

\bibitem[{Zhang et~al.(2019)Zhang, Kishore, Wu, Weinberger, and Artzi}]{zhang2019bertscore}
Tianyi Zhang, Varsha Kishore, Felix Wu, Kilian~Q Weinberger, and Yoav Artzi. 2019.
\newblock Bertscore: Evaluating text generation with bert.
\newblock \emph{arXiv preprint arXiv:1904.09675}.

\bibitem[{Zhao et~al.(2023)Zhao, Zhou, Li, Tang, Wang, Hou, Min, Zhang, Zhang, Dong et~al.}]{zhao2023survey}
Wayne~Xin Zhao, Kun Zhou, Junyi Li, Tianyi Tang, Xiaolei Wang, Yupeng Hou, Yingqian Min, Beichen Zhang, Junjie Zhang, Zican Dong, et~al. 2023.
\newblock A survey of large language models.
\newblock \emph{arXiv preprint arXiv:2303.18223}.

\bibitem[{Zheng et~al.(2023)Zheng, Chiang, Sheng, Zhuang, Wu, Zhuang, Lin, Li, Li, Xing et~al.}]{zheng2023judging}
Lianmin Zheng, Wei-Lin Chiang, Ying Sheng, Siyuan Zhuang, Zhanghao Wu, Yonghao Zhuang, Zi~Lin, Zhuohan Li, Dacheng Li, Eric Xing, et~al. 2023.
\newblock Judging llm-as-a-judge with mt-bench and chatbot arena.
\newblock \emph{arXiv preprint arXiv:2306.05685}.

\end{thebibliography}

\clearpage
\appendix
\section{Appendix}

\subsection{Dataset details}
\label{sec:dataset_details}
\begin{itemize} [leftmargin=10pt]
    \item {\bf Prime number}: we query the primality of 500 randomly chosen primes between 1,000 and 20,000; the factual answer is always \promptgpt{Yes}.  We synthesize hallucinated answers by \promptgpt{No, it is not a prime number.} and annotate the answers automatically. 

    \item {\bf Senator search}: the dataset consists of 500 questions that follow the template: \promptgpt{Was there ever a US senator that represented the state of $x$ and whose alma mater was $y$?} where $x$ is a U.S. state and $y$ is U.S. college. The correct answer is always \promptgpt{No}. We also generate hallucinated answers like \promptgpt{Yes, there was.} with annotation \promptgpt{False}.
    \item {\bf HotpotQA}: We randomly sample 250 examples from the training set of HotpotQA \cite{yang2018hotpotqa} and generate hallucinated answers drawn from \texttt{gpt-3.5-turbo}. Then we manually annotate the answers by comparing the ground truth and knowledge. 
    \item {\bf NQ-open-halu}: Natural Questions (NQ)-open \cite{lee2019latent} is a more challenging open domain QA benchmark \cite{kwiatkowski2019natural}. We use the same setting as HotpotQA-halu to create a small-scale dataset that consists of 250 non-factual and factual examples with manual annotations. 
\end{itemize}

\subsection{Prompt Set for \saccheck~and \sacsummary }
\label{sec:prompt_set}
Table \ref{tab:check} and \ref{tab:summary} provide the example of prompt used in \methodNamenew framework 
\begin{table}[ht!]
\centering

\caption{\saccheck~with \yopo}
\label{tab:check}

\begin{tabular}{p{8cm}}
\toprule
\textbf{Instruction}: \\
There are $M$ question-answering (QA) pairs, QA pair $1, 2, ..., M$. Please check if QA pair $i$ ($i=1,...,M$) is semantically equivalent to other QA pair $j$ ($j$ is not equal to $i$) and return how many QA pairs are semantically equivalent to QA pair $i$, only the number, no other words or explanation.

\textbf{For Example}:\\
QA pair 1: <the number of QA pairs are semantically equivalent to QA pair 1> \\
QA pair 2: <the number of QA pairs are semantically equivalent to QA pair 2> \\
... \\
QA pair $M$: <the number of QA pairs are semantically equivalent to QA pair $M$>  \\
\bottomrule
\end{tabular}
\end{table}

\begin{table}[ht!]
\centering
\caption{\sacsummary~with Generative Summary}
\label{tab:summary}

\begin{tabular}{p{8cm}}
\toprule
\textbf{Instruction}: \\
Given a specific question (or query) $Q$, there are $K$ semantically equivalent answers, i.e., Answer $1, 2,..., K$. Please summarize these answers into one improved and comprehensive answer.

\textbf{For Example}:\\
$Q$ : <Input question (or query)> \\
$A$ 1: <the semantically equivalent answer to the question $Q$>  \\
$A$ 2: <the semantically equivalent answer to the question $Q$>  \\
... \\
$A$ $K$: <the semantically equivalent answer to the question $Q$> \\
\bottomrule
\end{tabular}
\end{table}

\subsection{Discussion on the \sacsummary~Step}
\label{sec:summary}
We note that \sacsummary~is an optional step for further improving the final response quality if the budget (e.g., cost or latency) permits. If the budget is limited, the \sacsummary~step could be replaced with simply selecting the response from the candidate pool with the most votes.
Besides, one might be concerned that the involvement of an LLM in summarizing may bring additional bias to the final response. Although such bias exists in theory, the empirical observation from our experiments is that the bias brought by employing GPT-3.5 to summarize the response is almost negligible.

\end{document}